\documentclass[sigconf]{acmart}

\acmConference[AI-DEEDS 2026]{AI-DEEDS Workshop at ACM e-Energy}{June 22, 2026}{Banff, Canada}
\acmYear{2026}
\renewcommand\footnotetextcopyrightpermission[1]{}
\usepackage{booktabs}
\usepackage{graphicx}
\usepackage{amsmath}

\newcommand{\sys}{\texttt{sbsim}}

\begin{document}

\title{Quantifying the Energy Floor: Direct Measurement and Replay Buffer Bias in SAC-Based HVAC Control on \sys{}}

\author{Bo Li}
\affiliation{%
  \institution{Shanghai Jiao Tong University}
  \department{College of Smart Energy}
  \streetaddress{800 Dong Chuan Rd.}
  \city{Shanghai}
  \postcode{200240}
  \country{China}
}
\email{libob0709@sjtu.edu.cn}

\author{Chen Zhang}
\authornote{Corresponding author.}
\affiliation{%
  \institution{Shanghai Jiao Tong University}
  \department{College of Smart Energy}
  \streetaddress{800 Dong Chuan Rd.}
  \city{Shanghai}
  \postcode{200240}
  \country{China}
}
\email{chenzhang87@sjtu.edu.cn}

\begin{abstract}
We quantify the energy floor---the minimum achievable cost given action space constraints---for Soft Actor-Critic (SAC) HVAC control on the \sys{} calibrated building simulator.
Through minimum-action experiments, we directly measure this floor at \$35.51/day, dominated by continuous electrical loads (\$35.44, 99.8\%) with negligible gas consumption.
The standard SAC baseline, initialized with schedule-policy replay buffer transitions, converges to \$37.18/day---4.7\% above the floor.
We identify buffer initialization as the dominant source of sub-optimality in this scenario: training from an empty buffer reduces cost to \$35.57/day, eliminating 96\% of the gap.
Expanding the supply water temperature range by 10\,K yields negligible additional savings (\$0.03/day), and further expansion triggers physical constraint violations.
We additionally uncover a discount factor coupling ($\gamma_\text{eff} = 0.891$) shrinking the effective planning horizon from 8.3 h to 46 min—a benchmark-wide issue warranting audit.
Systematic ablation across planning horizon, reward weights, and observation enrichment confirms all pre-filled-buffer configurations cluster within 0.7\% (\$37.18--\$37.42), demonstrating that equipment minimum power---not algorithmic design---imposes the binding constraint.
\end{abstract}



\maketitle

\section{Introduction}
\label{sec:intro}

Buildings account for approximately 40\% of total energy consumption and carbon emissions in the United States, with HVAC systems responsible for the majority~\cite{doe2011}.
Traditional building management systems rely on fixed schedules that miss optimization opportunities arising from variable occupancy and building thermal inertia.
Reinforcement learning (RL) offers a framework for learning adaptive policies that balance comfort, energy cost, and carbon emissions.
Goldfeder and Sipple~\cite{goldfeder2024} introduced \sys{}, a calibrated digital twin of a real office building, and showed that a SAC agent trained for 4{,}000 iterations with Vizier hyperparameter search achieves an 8\% reward improvement over the rule-based schedule.

A natural question follows: can further algorithmic improvements---reward shaping, horizon extension, or richer observations---push costs below the SAC baseline?
Through systematic investigation, we find they cannot.
The building's action space imposes a structural \emph{energy floor}: the minimum achievable cost dictated by HVAC equipment that cannot be shut off.
In the SB1 summer scenario, this floor is almost entirely electrical.
We go beyond indirect inference and directly measure this floor via minimum-action experiments, revealing that the SAC baseline's sub-optimality stems primarily from replay buffer initialization bias rather than algorithmic limitations.

Our contributions are:
\textbf{(1)}~We directly measure the energy floor at \$35.51/day via minimum-action baselines, revealing that it is dominated by continuous electrical loads (\$35.44, 99.8\%) with negligible gas (\$0.07)---refining the conventional ``minimum equipment power'' intuition.
\textbf{(2)}~We identify replay buffer initialization as the dominant factor in the SAC baseline's sub-optimality, accounting for 96\% of the gap above the floor.
\textbf{(3)}~We document a discount factor coupling ($\gamma_\text{eff} = 0.891$) reducing the effective planning horizon to 46~minutes---a benchmark-wide configuration peculiarity warranting audit.
\textbf{(4)}~We confirm through systematic ablation that action space constraints---not algorithmic choices---are the binding bottleneck.

\section{Background}
\label{sec:background}

\subsection{Smart Building Simulator (\sys{})}

The \sys{} simulator~\cite{goldfeder2024} provides a calibrated digital twin of a two-story, 93{,}858\,ft$^2$ commercial office building (SB1) in Mountain View, California. The simulator instantiates 126 thermal zones, each serviced by a variable air volume (VAV) terminal, plus a shared air handling unit (AHU) and boiler (128 devices total).
The simulation runs at 5-minute timesteps; each episode spans one full day (288~steps).
Occupancy follows a stochastic arrival--departure model peaking at ${\sim}$125~persons between 15:00--19:00.

\paragraph{Action space.}
The agent controls two continuous setpoints normalized to $[-1, 1]$: supply air heating temperature (native range 285--300\,K) and supply water temperature (native range 310--355\,K).
Critically, both setpoints have \emph{fixed lower bounds}---the HVAC system cannot be turned off entirely, ensuring minimum energy expenditure even at the lowest settings.

\paragraph{Observation space.}
The default observation vector ($\mathbb{R}^{53}$) includes AHU and boiler sensor readings, histogram-reduced zone temperature distributions (19~bins), damper and airflow setpoints, and auxiliary temporal features (time-of-day and day-of-week as sine/cosine pairs).

\subsection{3C Regret Reward}
\label{sec:3c-reward}

The reward $r \in [-1, 0]$ combines three normalized components:
\begin{equation}
    r = w_p \cdot C_1 - w_e \cdot C_2 - w_c \cdot C_3
    \label{eq:3c-reward}
\end{equation}
where $C_1 \in [-1, 0]$ is comfort regret (zero when all \emph{occupied} zones are within the heating/cooling deadband; defined as $C_1 = 0$ during unoccupied periods regardless of temperature), $C_2 \in [0, 1]$ is normalized energy cost (ratio of actual to maximum capacity), and $C_3 \in [0, 1]$ is normalized carbon emission.
Default weights are $w_p = 0.2$, $w_e = 0.4$, $w_c = 0.4$.

\subsection{Soft Actor-Critic}

SAC~\cite{haarnoja2018} maximizes a maximum-entropy objective $J(\pi) = \sum_t \mathbb{E}[r_t + \alpha \mathcal{H}(\pi(\cdot|s_t))]$, where $\alpha$ is a learnable temperature.
Its off-policy nature and stability make it well-suited for building control, where each episode requires a computationally expensive physics simulation.
Following \cite{goldfeder2024}, we use SAC as the baseline RL algorithm.

\section{Method}
\label{sec:method}

\subsection{Baseline Replication}
\label{sec:baseline}

We train a standard SAC agent for 615 iterations (200 gradient updates per iteration, batch size 256) with the default 3C Regret weights.
The replay buffer (capacity 50K) is pre-populated with 1{,}440 transitions from the rule-based schedule policy, which operates at supply water 350\,K during occupied hours (06:00--19:00) and 315\,K otherwise.
The agent converges within ${\sim}$200 iterations, achieving a daily cost of \$37.18 with zero comfort violations.

\subsection{Direct Energy Floor Measurement}
\label{sec:energy-floor}

To establish the minimum achievable cost, we evaluate a deterministic \emph{minimum-action} policy that sets both action dimensions to $-1$ at every timestep, corresponding to the lowest available setpoints (supply air 285\,K, supply water 310\,K).
This policy represents the absolute lower bound achievable within the current action space, as it minimizes all controllable energy expenditure.
Even at minimum setpoints, continuous electrical loads (fan, pump) cannot be eliminated, establishing a non-zero cost floor whose magnitude we measure empirically.

\subsection{Expanded Action Space}
\label{sec:expanded}

We investigate whether extending the supply water temperature lower bound from 310\,K to 300\,K enables further cost reduction.
We evaluate both a minimum-action policy (fixed at 300\,K) and an SAC agent trained within this expanded range.
During preliminary experiments, setting the lower bound to 290\,K triggered a simulator physics constraint at timesteps when outdoor temperature exceeded the setpoint (e.g., \texttt{water\_temp 290.0\,K < outside\_temp 290.15\,K}), confirming that 310\,K is comfortably above the time-varying physical limit imposed by ambient conditions in the Mountain View summer climate (nighttime lows ${\sim}$290\,K / 17$^\circ$C, daytime highs ${\sim}$293\,K / 20$^\circ$C).

\subsection{Replay Buffer Bias Investigation}
\label{sec:buffer-bias}

The standard training protocol pre-populates the replay buffer with 1{,}440 transitions from the schedule policy, which operates at supply water 350\,K during occupied hours and 315\,K otherwise---far above the optimal minimum of 310\,K.
While replay buffer pre-filling with demonstration data is a standard technique in off-policy RL with documented benefits in many settings (e.g., DQfD~\cite{hester2018dqfd}, DDPGfD~\cite{vecerik2017ddpgfd}), we hypothesize that under the specific conditions of this benchmark---a substantially sub-optimal demonstration policy (${\sim}$16\% above the floor) combined with an optimal action at the action space boundary---the pre-filled transitions induce a persistent Q-function bias.
To test this, we train SAC agents with \emph{empty} replay buffers (both in the original 310--355\,K and expanded 300--355\,K action spaces), allowing the agent to learn exclusively from self-collected experience.

\subsection{Discount Factor Coupling}
\label{sec:discount}

We discover that \sys{} sets \texttt{discount\_factor\,=\,0.9}, emitted as the per-step discount in every non-terminal \texttt{TimeStep}.
TF-Agents' SAC multiplies this with $\gamma$ in the TD~target:
\begin{equation}
    y = r + \gamma \cdot d_{\text{env}} \cdot V(s') = r + 0.99 \times 0.9 \times V(s')
    \label{eq:gamma-mismatch}
\end{equation}
yielding $\gamma_\text{eff} = 0.891$---a horizon of only $1/(1{-}0.891) \approx 9$~steps (46\,min), too short to anticipate occupancy transitions or exploit the building's thermal inertia.

\subsection{Systematic Ablation}
\label{sec:ablation}

To confirm that the energy floor is robust across algorithmic choices, we evaluate seven additional configurations spanning three axes: (1)~reward shaping (time-adaptive occupancy-dependent weights, and energy-aggressive weights with $w_p\!=\!0.05$), (2)~planning horizon (correcting $d_\text{env}$ to 1.0), and (3)~observation enrichment (multi-frequency Fourier encoding, $\mathbb{R}^{53} \to \mathbb{R}^{61}$).
All ablation experiments use the schedule-policy pre-filled replay buffer.

\section{Experiments}
\label{sec:experiments}

\subsection{Setup}

All experiments use the default SB1 evaluation configuration shipped with the \sys{} repository, which sets the episode start to July~6, 2023 (mild summer in Mountain View: outdoor temperatures 15--20$^\circ$C). This date falls within the same season as the simulator calibration data of~\cite{goldfeder2024} (July~10--13, 2023), ensuring operation within the validated regime. We use 1-day episodes (288~steps).
SAC agents use two-layer actor and critic networks (256~units each), learning rate $3 \times 10^{-4}$, target update $\tau = 0.005$.
Agents train for 300--615 iterations (${\sim}$10--20\,h each on a single CPU), using segmented restarts every 15 iterations to manage memory growth.
We report daily cost (\$), daily CO$_2$ (kg), and comfort violations (\%).

\subsection{Results}

\begin{table}[t]
\centering
\caption{Performance on SB1 (1-day episodes). ``Buffer'' indicates replay buffer initialization. The energy floor (\$35.51) is directly measured via minimum-action experiments. Best cost in \textbf{bold}.}
\label{tab:results}
\small
\begin{tabular}{l l r r c}
\toprule
\textbf{Method} & \textbf{Buffer} & \textbf{Cost}$\downarrow$ & \textbf{CO$_2$}$\downarrow$ & \textbf{Cmf.} \\
\midrule
Schedule Baseline & --- & 41.27 & 22.82 & 0\% \\
\midrule
SAC Default & Pre-filled & 37.18 & 20.42 & 0\% \\
\;+ 7 ablations & Pre-filled & 37.35--37.42 & 20.47--20.80 & 0\% \\
\midrule
SAC (original space) & Empty & 35.57 & 19.76 & 0\% \\
SAC (expanded space) & Empty & 35.51 & 19.50 & 0\% \\
\midrule
Min Action (310\,K) & --- & 35.51 & 19.55 & 0\% \\
Min Action (300\,K) & --- & \textbf{35.48} & \textbf{19.27} & 0\% \\
\bottomrule
\end{tabular}
\end{table}

\begin{table}[t]
\centering
\caption{Energy floor gap decomposition. The \$1.67 gap between SAC Default and the energy floor is decomposed by source.}
\label{tab:gap}
\small
\begin{tabular}{l r r}
\toprule
\textbf{Component} & \textbf{Cost (\$)} & \textbf{\% of Gap} \\
\midrule
SAC Default (pre-filled buffer) & 37.18 & --- \\
\;\;$-$ Buffer bias & $-$1.61 & 96\% \\
SAC (empty buffer) & 35.57 & --- \\
\;\;$-$ Residual (entropy/exploration) & $-$0.06 & 4\% \\
\textbf{Energy Floor} (min-action) & \textbf{35.51} & --- \\
\bottomrule
\end{tabular}
\end{table}

\begin{figure}[t]
    \centering
    \includegraphics[width=\columnwidth]{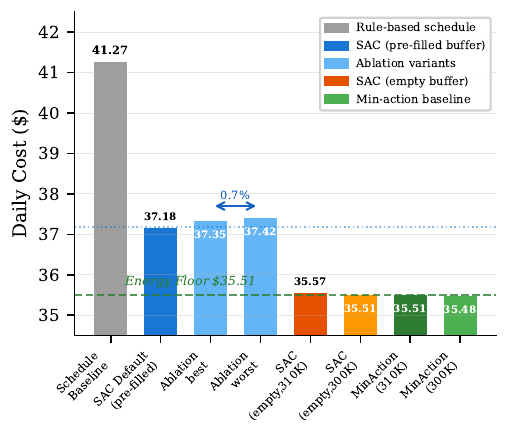}
    \caption{Daily energy cost across all methods. The dashed line marks the directly measured energy floor (\$35.51). Pre-filled-buffer SAC configurations cluster at \$37.18--\$37.42; empty-buffer SAC converges to \$35.57, near the floor. Total gap \$1.67; buffer bias accounts for \$1.61 (96\%), residual entropy/exploration \$0.06 (4\%).}
    \label{fig:energy-floor}
\end{figure}

\begin{figure}[t]
    \centering
    \includegraphics[width=\columnwidth]{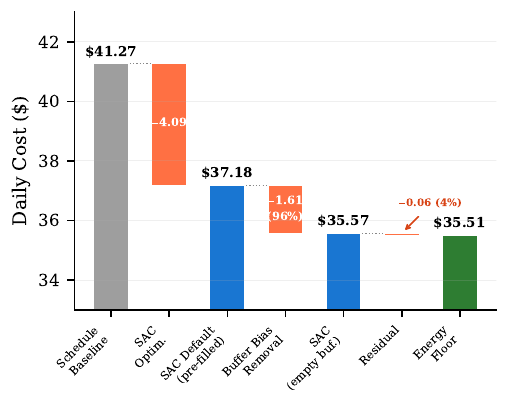}
    \caption{Gap decomposition from Schedule Baseline (\$41.27) to the energy floor (\$35.51). SAC optimization accounts for \$4.09; removing buffer bias recovers an additional \$1.61 (96\% of the remaining gap). Expanding the water temperature range by 10\,K yields only \$0.03.}
    \label{fig:gap-analysis}
\end{figure}

\begin{figure}[t]
    \centering
    \includegraphics[width=\columnwidth]{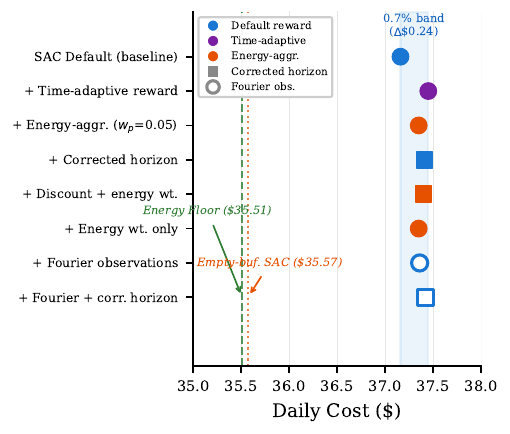}
    \caption{Ablation sensitivity analysis. Each point represents a configuration varying reward weights, planning horizon, or observation encoding. All pre-filled-buffer configurations fall within a 0.7\% band (\$37.18--\$37.42), far above the empty-buffer SAC (\$35.57) and the energy floor (\$35.51). Marker shape encodes horizon; fill encodes observation type; color encodes reward scheme.}
    \label{fig:ablation}
\end{figure}

Table~\ref{tab:results} presents the main results.
SAC Default reduces daily cost by 9.9\% (\$41.27$\to$\$37.18) and
carbon by 10.5\%, consistent with the ${\sim}$8\% return improvement
in~\cite{goldfeder2024} despite differing in episode length (1~day
vs.\ 2~days) and metric. Rapid convergence (${\sim}$200 iterations)
suggests a simple optimization landscape.

\paragraph{Ablation sensitivity.}
Seven additional configurations systematically vary three axes: (i)~\emph{reward shaping}---time-adaptive occupancy-dependent weights versus energy-aggressive weights ($w_p\!=\!0.05$, $w_e\!=\!0.55$, $w_c\!=\!0.40$); (ii)~\emph{planning horizon}---correcting the discount to $\gamma_\text{eff} = 0.99$; and (iii)~\emph{observation enrichment}---multi-frequency Fourier encoding ($\mathbb{R}^{53} \!\to\! \mathbb{R}^{61}$).
As shown in Figure~\ref{fig:ablation}, all pre-filled-buffer configurations fall within a 0.7\% cost band (\$37.18--\$37.42).
This insensitivity is striking: techniques that typically yield improvements in other RL domains produce negligible effect here.
The tight clustering persists under pairwise and triple combinations (e.g., Fourier + corrected horizon: \$37.42), indicating no synergistic interaction among the three axes.

\paragraph{Energy floor.}
The minimum-action policy at 310\,K achieves a daily cost of \$35.51.
Cost decomposition reveals this is almost entirely electrical (\$35.44, 99.8\%) with negligible gas (\$0.07).
Extending the water temperature lower bound to 300\,K yields only \$0.03 additional savings (\$35.48); this reduction comes exclusively from the gas component (\$0.07 $\to$ \$0.03), while electrical cost is unchanged.
This confirms that continuous electrical loads---not the temperature setpoint range---are the binding constraint.
A~further extension to 290\,K triggers the simulator physics constraint \texttt{water\_temp < outside\_temp} at timesteps when ambient temperature exceeds the setpoint, confirming a time-varying thermodynamic lower bound near ${\sim}$290\,K.
Notably, the minimum-action policy achieves zero comfort violations: in this mild climate, the building's thermal mass and solar gains maintain zone temperatures within the comfort deadband without active heating.

\paragraph{Buffer bias.}
Table~\ref{tab:gap} decomposes the \$1.67 gap between SAC Default and the energy floor.
Training SAC with an empty buffer---in the \emph{same} 310--355\,K action space---reduces cost to \$35.57, eliminating 96\% of the gap (Figure~\ref{fig:gap-analysis}).
The pre-filled buffer, containing 1{,}440 transitions at supply water 350/315\,K (up to 40\,K above the optimal minimum), biases the Q-function toward high-temperature actions.
SAC's entropy regularization further inhibits convergence to the deterministic minimum-action optimum: the maximum-entropy objective penalizes the near-deterministic policy required to consistently select both action dimensions at $-1$.
This also explains why all eight pre-filled-buffer configurations cluster near \$37.2: the shared biased initialization dominates any algorithmic variation.

\paragraph{Discount factor coupling.}
Correcting the environment discount to $d_\text{env} = 1.0$ (restoring $\gamma_\text{eff} = 0.99$) yields only marginal cost changes (\$37.41 vs.\ \$37.18 under default weights).
However, under comfort-weighted rewards ($w_p = 0.4$), the extended horizon enables the agent to pre-heat aggressively, \emph{increasing} cost by 3.6\% (\$38.49).
This asymmetry reveals that horizon extension amplifies whatever objective the reward encodes rather than universally improving efficiency.
The 46-minute effective horizon under $\gamma_\text{eff} = 0.891$ limits the agent's ability to exploit thermal inertia for predictive pre-conditioning strategies.

\section{Analysis}
\label{sec:analysis}

\paragraph{Physical interpretation.}
The cost decomposition in Section~\ref{sec:experiments} reveals that the energy floor is overwhelmingly electrical (\$35.44 of \$35.51)---a finding with important physical implications.
Applying the PG\&E commercial tariff (${\sim}$\$0.16--0.20/kWh) implies an average continuous electrical load of ${\sim}$7--9\,kW over 24~hours.
This is substantially lower than equipment nameplate ratings (AHU fan rated ${\sim}$30\,kW; boiler pump ${\sim}$5\,kW), because at minimum action the equipment operates at ${\sim}$20--25\% of rated capacity.
This refines the ``minimum operating power'' intuition: the floor is determined not by what equipment \emph{could} draw, but by what it \emph{actually} draws under continuous operation at lowest setpoints.
The minimal gas component reflects the mild climate: VAV reheat coils at 310\,K supply sufficient heat through air mass flow alone, keeping zones within the comfort deadband (294--297\,K during occupied hours) without significant boiler activation.
Lowering the supply water bound thus primarily reduces the already-tiny gas component, explaining the negligible \$0.03 savings from expanded action space.
The minimum-action policy simultaneously achieves zero comfort violations and minimum cost---it is the scenario-specific optimal policy.

\paragraph{Buffer bias mechanism.}
The schedule policy pre-fills the buffer with transitions at supply water 350\,K (77$^\circ$C) during occupied hours---40\,K above the optimal minimum.
In this mild climate, even 310\,K water through VAV reheat coils produces sufficient heating.
The 350\,K operation represents substantial over-heating, wasting energy on unnecessary boiler output and pump work.
The Q-function, trained initially on these high-cost transitions, develops a landscape biased toward high temperatures.
While off-policy SAC can theoretically overcome this bias given sufficient exploration, the combination of a pre-warmed Q-function and entropy regularization slows convergence toward the deterministic low-action optimum.
SAC parameterizes its policy as a squashed Gaussian $a = \tanh(\mathcal{N}(\mu, \sigma^2))$, for which the log-density diverges as $a \to \pm 1$; concentrating policy mass at the action boundary drives entropy toward $-\infty$, which the learnable temperature $\alpha$ counteracts by inflating the entropy bonus.
This structural opposition to boundary policies is independent of buffer contents and provides a plausible mechanism for the \$0.06 residual gap between empty-buffer SAC (\$35.57) and the directly measured floor (\$35.51).
The tight clustering of all eight pre-filled-buffer configurations near \$37.2 provides empirical evidence that the buffer bias is the common limiting factor.

\paragraph{When does buffer bias matter?}
Our findings should not be interpreted as a blanket critique of buffer pre-filling, which has demonstrated benefits in many RL applications~\cite{hester2018dqfd,vecerik2017ddpgfd}.
Rather, three conditions appear necessary for pre-filling to become the dominant source of sub-optimality:
\textbf{(i)~Demonstration sub-optimality.}
When the demonstration policy is far from optimal (here, ${\sim}$16\% above the floor), the Q-function inherits this gap.
In settings where demonstrations approximate the optimal policy, pre-filling accelerates convergence rather than impeding it.
\textbf{(ii)~Boundary optimum.}
When the optimal policy lies at the action space boundary (here, action~$= [-1, -1]$), SAC's maximum-entropy objective structurally opposes convergence to deterministic actions; pre-filled transitions from a non-boundary policy compound this difficulty.
\textbf{(iii)~Comfort non-binding.}
When external constraints (comfort, safety) do not force action into the interior, the gap between demonstration and optimum can become large.
In binding-comfort scenarios (e.g., winter heating), demonstration and optimum likely converge in the action space interior, reducing the bias.
The SB1 summer scenario evaluated here satisfies all three conditions simultaneously, explaining the dramatic effect (\$1.61, 96\% of the gap).

\paragraph{Implications for reward design.}
The 3C Regret defines $C_1 = 0$ whenever zones are unoccupied, regardless of temperature.
This means reducing comfort weight $w_p$ during unoccupied hours is a no-op: it does not change the reward experienced by the agent.
The energy and carbon components already dominate ($w_e + w_c = 0.8$), so the agent is maximally incentivized to save energy during unoccupied hours under the default weights.
The flat energy profile observed across all hours reflects the action space constraint rather than a sub-optimal policy.

\paragraph{Generalizability.}
The energy floor finding is scoped to scenarios satisfying the three conditions identified above.
Quantitatively, our results hold while the outdoor--setpoint gap remains within ${\sim}$5--10\,K (here, 1--6\,K in Mountain View summer).
In colder climates where this gap exceeds 15--20\,K, active boiler operation becomes mandatory to maintain comfort, raising the floor substantially and shifting the optimal action away from the boundary into the action-space interior.
This restores the conditions under which standard SAC---with entropy regularization and demonstration pre-filling---is well-suited, and genuine algorithmic improvements likely become measurable.
Buildings with variable-speed fans or on/off control would similarly lower the floor, creating room for algorithmic gains.
Our results thus characterize a \emph{ceiling} on RL performance in the SB1 mild-climate single-day scenario, not a universal limitation.
The recently extended \sys{} suite~\cite{goldfeder2024suite} introduces 11~buildings spanning diverse climates, management systems, and sizes, along with a Physically Informed Neural Network (PINN) building model as a simulator alternative---providing a natural testbed for validating the energy floor phenomenon across building types and climates.

\section{Related Work}
\label{sec:related}

Deep RL for HVAC control has progressed from tabular Q-learning~\cite{wei2017} to continuous-action algorithms on physics-based simulators~\cite{zhang2018, biemann2021}.
Model-based approaches leveraging differentiable dynamics~\cite{drgona2020, chen2019} offer sample efficiency but require accurate thermal models.
The \sys{} simulator~\cite{goldfeder2024} provides a calibrated benchmark with realistic occupancy and pricing; our work extends its evaluation with direct energy floor measurement.
Henderson et~al.~\cite{henderson2018} showed that many reported RL improvements fail rigorous evaluation; our ablation contributes similar evidence for HVAC, identifying action space constraints as the binding bottleneck~\cite{dulac2021}.
Replay buffer pre-filling with demonstrations has shown clear benefits in robotics and game domains~\cite{hester2018dqfd,vecerik2017ddpgfd,de2018experience}; we provide a concrete HVAC case where pre-filling with a \emph{substantially sub-optimal} policy inflates converged costs by 4.7\%, illustrating that demonstration quality relative to the optimum is a critical factor.

\section{Conclusion}
\label{sec:conclusion}

We directly measured the energy floor for SAC-based HVAC control on \sys{} at \$35.51/day and identified four key findings:

\textbf{(1) Replay buffer bias.}
Under the conditions evaluated here, pre-filling the buffer with sub-optimal schedule-policy transitions inflates the converged cost to \$37.18 (4.7\% above floor).
Training from an empty buffer achieves \$35.57, eliminating 96\% of the gap---indicating that SAC itself can find near-optimal solutions in this scenario when not constrained by biased initialization.
This finding pertains to scenarios where the demonstration policy is substantially sub-optimal and the optimal action lies at the action space boundary; it does not generalize to settings where pre-filling with near-optimal demonstrations is known to be beneficial.

\textbf{(2) Action space as binding constraint.}
Expanding the water temperature range by 10\,K saves only \$0.03---exclusively from gas reduction, while the dominant electrical component (\$35.44/day, 99.8\%) is invariant to setpoint changes; systematic ablation across planning horizon, reward weights, and observations yields costs within 0.7\% (\$37.18--\$37.42).
The floor is imposed by continuous electrical loads (fan, pump), not by algorithmic design.

\textbf{(3) Discount coupling.}
The environment's $d_\text{env} = 0.9$ compounds with $\gamma = 0.99$, yielding $\gamma_\text{eff} = 0.891$ (46-min horizon)---a benchmark-wide configuration issue.

\textbf{(4) Algorithmic insensitivity.}
Systematic ablation across reward shaping, planning horizon, and observation enrichment confirms action space constraints—not algorithmic choices—as the binding bottleneck.

We recommend reporting minimum-action baselines, auditing demonstration quality before buffer pre-filling, and checking discount coupling. Further gains require on/off control or variable-speed fans.

\begin{acks}
We thank Judah Goldfeder and John Sipple for open-sourcing sbsim and for helpful workshop correspondence. We acknowledge the extended \sys{} suite (Goldfeder et al., 2025) for inspiring future multi-building evaluation.
\end{acks}

\bibliographystyle{ACM-Reference-Format}

\end{document}